%% file: main.tex
\documentclass[10pt,twocolumn,letterpaper]{article}

\usepackage{wacv}              % To produce the CAMERA-READY version
\usepackage{graphicx}
\usepackage{amsmath}
\usepackage{amssymb}
\usepackage{booktabs}
\usepackage{tabularray}
\usepackage{array}

\usepackage[pagebackref,breaklinks,colorlinks]{hyperref}

\usepackage[capitalize]{cleveref}
\crefname{section}{Sec.}{Secs.}
\Crefname{section}{Section}{Sections}
\Crefname{table}{Table}{Tables}
\crefname{table}{Tab.}{Tabs.}

\def\wacvPaperID{3} % *** Enter the WACV Paper ID here
\def\confName{WACV}
\def\confYear{2025}

\begin{document}

\title{SEER-ZSL: Semantic Encoder-Enhanced Representations \\ for Generalized Zero-Shot Learning}

\author{William Heyden \vspace{-7em}
\and
Habib Ullah \vspace{-7em}
\and
Muhammad Salman Siddiqui \vspace{-7em}
\and
Fadi Al Machot
\\Faculty of Science and Technology (REALTEK),\\Norwegian University of Life Sciences, 1433 Ås, Norway\\{\tt\small \{william.heyden, habib.ullah, muhammad.salman.siddiqui, fadi.al.machot\}@nmbu.no}
\\[-1.0ex]
}
\maketitle

%%%%%%%%% ABSTRACT
\begin{abstract}
    Zero-Shot Learning (ZSL) presents the challenge of identifying categories not seen during training. This task is crucial in domains where it is costly, prohibited, or simply not feasible to collect training data. ZSL depends on a mapping between the visual space and available semantic information. Prior works learn a mapping between spaces that can be exploited during inference. 
    We contend, however, that the disparity between meticulously curated semantic spaces and the inherently noisy nature of real-world data remains a substantial and unresolved challenge.
    In this paper, we address this by introducing a Semantic Encoder-Enhanced Representations for Zero-Shot Learning (SEER-ZSL). We propose a hybrid strategy to address the generalization gap. First, we aim to distill meaningful semantic information using a probabilistic encoder, enhancing the semantic consistency and robustness.
    Second, we distill the visual space by exploiting the learned data distribution through an adversarially trained generator. 
    Finally, we align the distilled information, enabling a mapping of unseen categories onto the true data manifold. We demonstrate empirically that this approach yields a model that outperforms the state-of-the-art benchmarks in terms of both generalization and benchmarks across diverse settings with small, medium, and large datasets. The complete code is available on GitHub.
    \footnote[2]{https://github.com/william-heyden/SEER-ZeroShotLearning}
\vspace{-.2cm}
\end{abstract}

\input{sections/introduction}
\input{sections/related_works}

\input{sections/methodology}
\input{sections/dataset}
\input{sections/experiments}

\input{sections/conclusion}

%%%%%%%%% REFERENCES
{\small
\bibliographystyle{ieee_fullname}
\bibliography{egbib}
}

\end{document}

%% file: sections/introduction.tex
\section{Introduction}
\label{sec:intro}

Zero-shot learning (ZSL) represents a fundamental challenge at the intersection of machine learning and computer vision. It addresses the problem of recognizing and classifying objects never seen during training. Recognizing unseen objects is highly relevant across a range of practical scenarios. For instance, in toddler and infant care, it could facilitate the identification of early signs of rare developmental disorders or previously unobserved behavioral anomalies. Similarly, in elderly care, this capability can aid in detecting unfamiliar patterns of movement or behavior that signal unanticipated health emergencies, such as a stroke or risks of falling. Beyond healthcare, this extends to identifying unknown species, discerning unfamiliar material compositions, or addressing limitations in data collection, emphasizing the broad utility of recognizing unseen classes across various domains \cite{rezaei2020zero}. Zero-shot learning accomplishes this through a mapping of shared semantic knowledge of the domain, in order to infer previously unknown areas of the data manifold \cite{ding2018generative}. Hence, it is evident that the compatibility between the shared knowledge and the primary domain plays a pivotal role. 

\begin{figure}[t]
    \centering
    \includegraphics[scale=0.4]{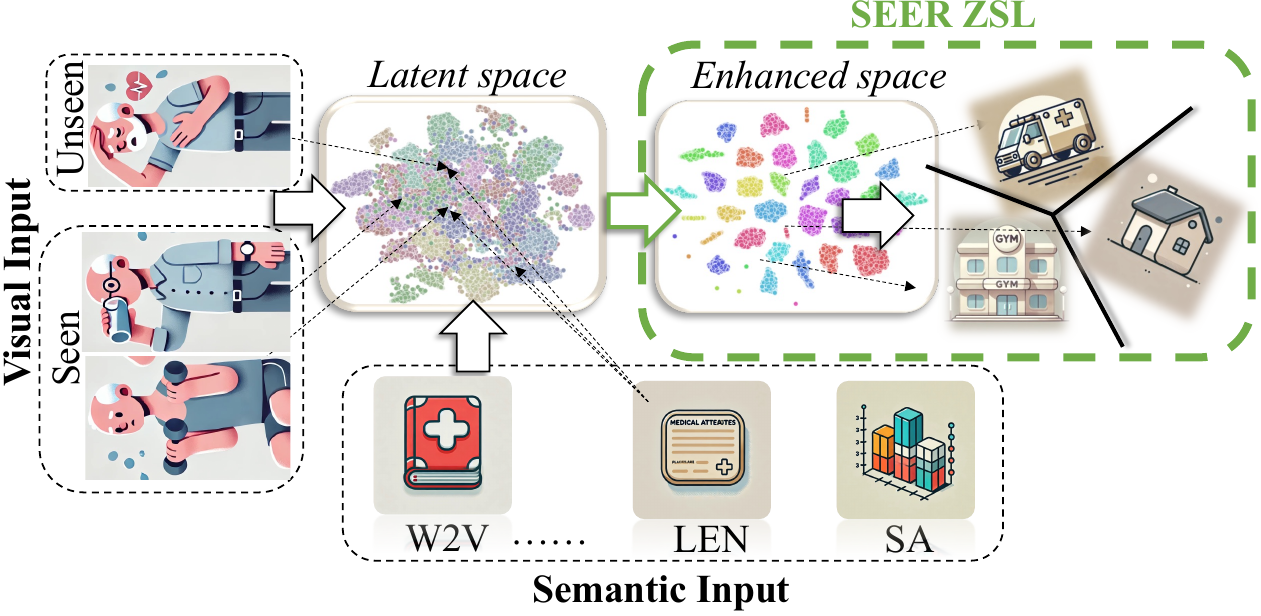}
    \caption{\small {Our Semantic Encoder-Enhanced Representation (SEER) model is specifically designed to address the challenges posed by inherent variances in semantic data, by effectively mapping auxiliary information to corresponding visual cues.
    %SEER enables accurate classification. 
    This capability is particularly valuable in applications like the visual monitoring of elderly individuals, where the model could recognize even unseen urgent situations (e.g., detecting signs of a heart attack) and suggest appropriate actions, such as alerting emergency services.
    This framework demonstrates the potential of SEER to extend beyond traditional action recognition tasks, facilitating the reliable detection of critical, unobserved scenarios.}}
    \label{fig:concept}
    \vspace{-.2cm}
\end{figure}

Traditional methods relied on matching visual attributes to object categories. This often led to a semantic gap \cite{lampert2009learning}. Modern approaches leverage deep learning models to mitigate this gap by learning a joint embedding space of visual and semantic features \cite{chen2020rethinking, mishra2018generative, chen2021semantics, frome2013devise, socher2013zero}. Building on this, the advent of latent space techniques in ZSL has received significant interest. These approaches allow for explicit approximation of data distributions, addressing the challenges associated with the semantic gap and improving generalization \cite{verma2017simple}.  
Recent studies have demonstrated latent representations' ability to bridge the gap between seen and unseen classes \cite{chen2021free, han2021contrastive, jiang2019transferable}.

We argue that there exists a disparity between how shared knowledge is implemented in benchmark settings and what can be expected in realistic scenarios \cite{rezaei2020zero}. The commonly used shared knowledge consists of semantic descriptions covering both seen and unseen categories. However, these semantic spaces are often meticulously crafted for specific purposes, posing limitations on models when applied in different, realistic scenarios \cite{xian2018zero}.
To address this limitation, our paper introduces a novel approach emphasizing the role of generalization capabilities in generative-based zero-shot learning. The central tenet of this approach revolves around achieving fine-grained control over semantic information while mitigating the unintended presence of semantic-unrelated information for the domain, as illustrated in Fig. \ref{fig:concept}. By effectively guiding the generative model to augment the existing knowledge, we can interpolate to unseen objects within the true data manifold. This hybrid strategy enhances the differentiation between categories and addresses the core problems of bias and hubness that have hindered previous ZSL methodologies \cite{lazaridou2015hubness}. Hence, our approach not only alleviates the challenges of dispersing semantic knowledge but also enhances the generalization capabilities necessary for advanced ZSL applications.
%, thus setting new benchmarks for AI systems in recognizing unseen objects. 
To summarize, we introduce three major contributions in this paper:

\begin{itemize}
    \item \textbf{Improved semantic control.} Generating latent features closely aligned with semantic attributes ensures enhanced class differentiation, which is crucial for class inference.
    %\item \textbf{Noise Mitigation:} Our method generates cleaner, class-specific features by focusing on noise filtering, leading to more accurate classification of unseen data.
    \item \textbf{Generalization and noise mitigation.} Enhanced control over the feature space allows our model to overcome biases towards seen classes and address the hubness problem.
    \item \textbf{Pseudo prototype generation.} The precise semantic and visual alignment enables accurate prototype generation of unseen objects.
    \item \textbf{Domain adaptation.} The model adapts to a variety of semantic sources through hyperparameters. This control allows for a universal adoption across datasets.
\end{itemize}

\subsection{Small Data Statement}
\label{small_data}
In this paper, we leverage widely recognized ZSL datasets exclusively to facilitate direct and fair comparisons with state-of-the-art models. Zero-shot learning aims to enable models to generalize and transfer knowledge to previously unseen categories; thus, we use classification accuracy as a proxy to evaluate this capability. Therefore, to validate and highlight the effectiveness of our proposed model, we utilize datasets commonly used in ZSL research. These datasets feature standard splits of seen and unseen training examples for a comparable assessment of transferability. Each dataset exhibits unique characteristics in terms of detail, size, and variability, which provides a diverse testing ground. As a result, our analysis is transferable into other domains. This could be rare categories, such as early-onset developmental disorders in infants or previously unknown, emerging categories such as patterns of mobility decline in the elderly. Nonetheless, our approach is designed to be dataset-agnostic. The application to additional datasets, though straightforward, lies beyond the scope of this work.

%The objective of zero-shot learning is to enable models to generalize knowledge to previously unseen categories without explicit training examples. This capability is crucial for scenarios where labeled data is scarce or unavailable. This could include: (I) rare categories, such as early-onset developmental disorders in infants, (II) previously unknown or emerging categories, such as novel patterns of mobility decline in the elderly, and (III) new compositions, such as recognizing unique combinations of social interactions or motor behaviors in toddlers. These applications highlight the potential of ZSL in addressing real-world challenges across diverse and impactful domains. In this paper, we utilize widely recognized benchmark datasets exclusively to facilitate direct comparison with state-of-the-art models. However, our approach is designed to be dataset-agnostic, making its extension to other datasets straightforward.

%% file: sections/related_works.tex
\section{Related work}
\label{sec:related}

\begin{figure*}[t]
    \centering
    \includegraphics[scale=0.45]{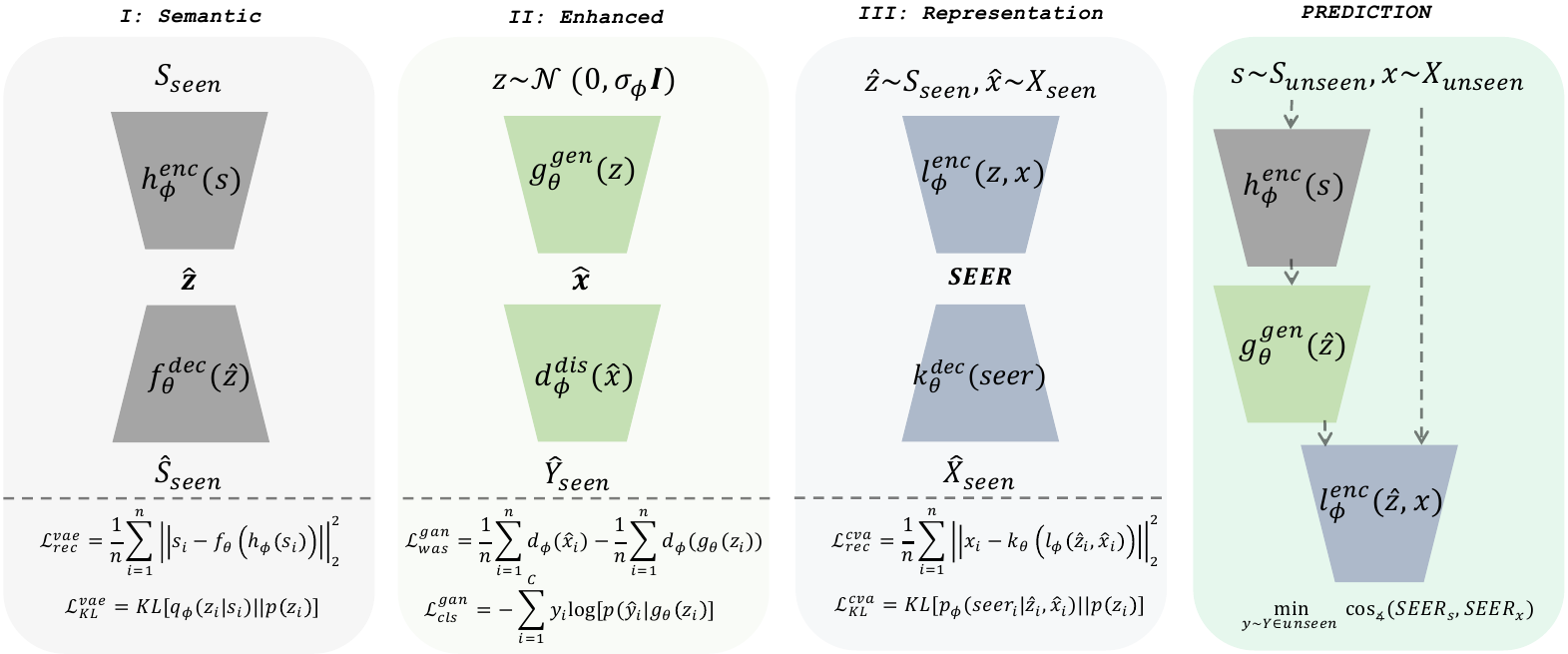}
    \caption{\small{We simultaneously train (I) an encoder to extract unbiased semantic information while prompting a generator to utilize the learned data distribution derived from the unknown probability space. (II) To guide the generator, we introduce a classification loss that encourages the development of high-level discriminative features. (III) Subsequently, we optimize a conditional variational encoder to align the extracted knowledge from the latent data distribution with the true distribution. This methodology enables classification in a SEER space using a simple interpretable distance metric.}}
    \label{fig:architecture}
    \vspace{-.3cm}
\end{figure*}

%In zero-shot learning (ZSL), a major focus has been on enhancing the latent space, which is essential for developing models for generalized zero-shot learning (GZSL) to classify both seen (trained on) and unseen (not trained on) classes. This section surveys the latest methods in ZSL, categorizing them based on their approach to latent space manipulation. 

We survey the latest methods by categorizing them based on their approaches to the latent space manipulation. 

\textbf{Contrastive learning approaches.} These methods enhance latent space learning by contrasting data with semantic embeddings, thus improving the link between seen and unseen classes. Examples include the Transferable Contrastive Network \cite{jiang2019transferable}, which ensures consistent visual-semantic alignment through knowledge transfer, and the hybrid GZSL framework \cite{han2021contrastive}, which combines generative and embedding models for improved feature mapping. Alternative approaches to align seen and unseen data are presented in \cite{fu2015transductive, sung2018learning}. Here, the authors introduce a calibration network and a relation network, respectively, aimed at learning to adjust for the uncertainty of unseen classes. However, generalizing to unseen classes remains a key challenge, as these methods need to adapt learned representations without bias \cite{akata2013label, fu2015zero, zhang2015zero, jiang2017learning}. Achieving a balance in visual-semantic alignment is also challenging, requiring careful hyperparameter calibration \cite{heyden2023integral}.

\textbf{Disentanglement-based methods.} Techniques like Generative-Based Disentanglement \cite{kazemi2019style} and Conditional VAE \cite{sohn2015learning} with a Total Correlation (TC) module focus on separating overlapping factors in the latent space for clearer, more transferable representations \cite{li2021generalized, chen2021semantics}. By leveraging the concept of decoupling \cite{almahairi2018augmented, lee2018diverse}, authors attempt to transfer learned domain-invariant features into distinct attribute vectors. Conversely, the authors \cite{higgins2016beta} adopt a straightforward KL-restriction technique to induce sparsity within the learned feature set. These methods aim to isolate meaningful features in data but sometimes struggle to filter out noise properly, leading to less accurate generalization to unseen classes \cite{bengio2013representation}. 
 
\textbf{Attention mechanisms and hierarchical models.} Attention-based methods, most notably in compositional ZSL, employ strategies like cross-attention for targeted attribute analysis \cite{hao2023learning}, self-attention to exploit interdependent structures \cite{khan2023learning} or attention pooling for attribute patches \cite{ruis2021independent}. Hierarchical models \cite{tong2019hierarchical, kim2023hierarchical, purushwalkam2019task} improve feature separation through layered disentanglement. Attention mechanisms and hierarchical models in zero-shot learning face challenges such as attribute imbalance, difficulty in transferring non-semantic features, and reliance on high-quality, detailed attributes. These issues can impact their effectiveness and generalization to unseen classes.

\textbf{Cross-modal and optimization techniques.} Models that integrate cross-modal latent spaces and latent space optimization techniques focus on aligning attributes across different modalities and enhancing generalization to unseen classes \cite{chen2023duet, min2020domain, liu2021goal, chen2022msdn, chen2022transzero}. Cross-modal and optimization techniques involve integrating latent spaces across different modalities and optimizing these spaces. Maintaining consistent attribute representation and ensuring that attributes are uniformly linked and accurately represented in each modality remains a core challenge of these methods \cite{zhao2019large, wang2021improving}.

%% file: sections/methodology.tex
\section{Methodology}
\label{sec:methods}

\textbf{Problem setting.} Zero-shot learning aims to develop a model capable of understanding visual and semantic indicators to classify unseen categories. Essentially, ZSL becomes necessary when labeled examples are unavailable for some classes in question. Thus, we can see our dataset as being divided into a training set with seen classes denoted as $D^{seen}$ with data $x \in X^n$, semantic information $s \in S^n$, and the labels $y \in Y^n$ for $n$ samples. $D^{unseen}$ represents the testing set with unseen data $x \in X^u$, semantic information $s \in S^u$, and the labels $y \in Y^u$ for $u$ samples. It is critical to ensure that $\mathrm{Y}^{n} \cap \mathrm{Y}^{u} = \phi$ and $S^n \cap S^u=S$, i.e., the category space is disjoint, but the semantics are shared. In generalised ZSL (GZSL), the testing domains contain samples from $D_{test} \in D^n \cap D^u$ \cite{pourpanah2022review}. The challenge lies in constructing a function $\mathbb{R}^{x} \cap \mathbb{R}^{s}  \rightarrow \mathrm{Y}$, which learns to predict the class category by learning how to extract knowledge from the training set which can be transferred to seen and unseen categories. Our ZSL method aims to mimic human adaptability to novel scenarios by leveraging models that can extract a latent representation of available modalities without prior examples. Our goal is to align these two spaces within a third manifold, enabling optimal separation of unseen classes. \newline

\textbf{(I) Semantics.} Our first step is to extract a lower-dimensional latent representation of the semantic space, sub-figure (a) in Fig. \ref{fig:architecture}. This process ensures the robustness and consistency of the semantic space representation. We first train a Variational Autoencoder (VAE) \cite{chen2016variational} that estimates a lower bound of the log-likelihood of the semantic space $\mathbb{E}_{z\sim q_{\phi}(z|s)} \left[ \log p_{\theta}(s|z) \right]$. The VAE is trained with the standard reconstruction loss and a Kullback-Leibler (KL) divergence loss:

\vspace{-.3cm}
\begin{equation}
    \begin{split}
    \mathcal{L}_{\text{VAE}}(h,f) = -\sum_{i=1}^{n} \mathbb{E}_{z_i \sim q_{\phi}(z_i|s_i)} \left[ \log p_{\theta}(s_i|z_i) \right] - \\ \beta\text{KL}(q_{\phi}(z_i|s_i) \| p(z_i))
    %\mathcal{L}_{\text{VAE}} = \mathcal{L}_{\text{recon}}(s, \hat{s}) + \beta %\mathcal{L}_{\text{KL}}(\mathcal{N}(0, I))
    \end{split}
    \label{LVAE}
\end{equation}

where the first term is the reconstruction loss, the second term is the KL divergence loss, and $z$ is the latent variable. The pseudo semantic space $\hat{s} \sim p_{\theta}(s|Z) $ is the conditional model density, and $\beta$ is a weighting coefficient. Exploiting the reconstruction loss enables coverage of high-density structures of the training data distribution while the KL loss imposes diversity constraints on the variability within the semantic space \cite{prokhorov2019importance}. 
%Furthermore, the KL loss imposes constraints on the spread of the generated latent space.
The loss of Eq. \ref{LVAE} ensures that the latent representation describes a joint distribution of the original semantics while retaining key characteristics of the true training distribution $p(s,z) \sim \mathbb{P}_r$.
In accordance with the domain adoption theory \cite{sanodiya2022manifold} we expect the semantic space $S$ to be closely related to the underlying true data manifold $\mathbb{P}_r$ of the joint space $\mathbb{E}_{\mathbb{P}(x,s)}[f(x,s)]=\int x\int [f(x,s)p(s|x)] dx$. By introducing the adjustable parameter $\beta$, we can balance the latent capacity for capturing information from the true data manifold distribution $\mathbb{P}_r$. For a continuous semantic distribution, the encoder parameterized by $\theta$ can be exploited to retain certain properties $\mathbb{E}_{z \sim \mathbb{P}_r}[p_{\theta}(z|s)]$ where feature-wise errors are minimized \cite{larsen2016autoencoding}. Assuming a continuous reconstructed space, the decoder $\hat{s} \sim q_{\theta}(s|z)$ captures the data distribution \cite{hou2017deep}, while offering an invariance to manipulation, often referred to as \textit{style} error \cite{gatys2015neural}. \newline

\textbf{(II) Enhanced.} Next, we utilize a Wasserstein Generative Adversarial Network (WGAN) \cite{arjovsky2017wasserstein} to capture the true training distribution of the visual space $x$. This is shown in Fig. \ref{fig:architecture}(b). By learning a Lipschitz continuous function $G$ from the support of the manifold of the latent representation $\mathbb{E}_{x \sim G(z)}[log D(x)]$. The generator $G$ is trained with a Wasserstein loss with gradient penalty in Eq. \ref{LGAN}. The first two terms are the objective of the generator and the critic, respectively, while the last term is the gradient penalty, regulated with $\alpha$, to enforce Lipschitz constraint without exploding the gradients \cite{gulrajani2017improved}.

\vspace{-.3cm}
\begin{equation}
    \begin{split}
    \mathcal{L}_{\text{WGAN}}(g, d) =  \frac{1}{n}\sum_{i=1}^{n} \mathbb{E}_{x_i \sim G_{z}}[D(x_i)] 
    \\ - \frac{1}{n}\sum_{i=1}^{n} \mathbb{E}_{x_i \sim \mathbb{P}_r}[D(x_i)] 
    \\ + \alpha \mathbb{E}_{\hat{x} \sim \mathbb{P}_{\hat{x}}} [(| \nabla_{\hat{x}} D(\hat{x})|_2-1)^2]
    %\mathcal{L}_{\text{WGAN}} = \mathcal{L}_{\text{Wasserstein-GP}}(g(z), X_{\text{seen}}) + \eta %L_{\text{cls}}
    \end{split}
    \label{LGAN}
\end{equation}

% $L_{\text{cls}}$ is defined as:

Let $G: z \rightarrow \hat{x}$ be a function of nonlinear affine transformation, then $G$ sustains the prior $z$ to be contained in $\mathbb{P}_r$ \cite{arjovsky2017towards}. 
To address the instability of the critic caused by potential disjoint support when $dim(z)<<dim(X)$ \cite{narayanan2010sample}, we propose an additional criterion aiming to partially align the two manifolds. By conditioning on a class-label loss $L_{\text{cls}}$ defined in Eq. \ref{LCLS}, $G(z)$ captures high-level features of the data.
Moreover, the manifold hypothesis \cite{narayanan2010sample} enables us to integrate lower-dimensional semantic information directly into the generation process, without the loss of generalization. This approach maintains the joint estimation of the distribution. The critic $D$ aims to distinguish between the generated $\hat{x}$ and real samples $x$.
%within a specific segment.

%\begin{figure}
%    \centering
%    \includegraphics[scale=0.40]{figures/methods_exp.pdf}
%    \caption{Methodology}
%    \label{fig:method}
%\end{figure}

\begin{equation}
    L_{\text{cls}} = -\sum_{n}\sum_{C} \log[Pr(Y_C^n|\hat{x}^n)]
    \label{LCLS}
\end{equation}

By guiding the generator with classification loss, we mitigate the critic $D$, ensuring more manageable gradients for the generator to enhance its performance. The overall loss for the Semantic-Enhanced embedding is:

\vspace{-.3cm}
\begin{equation}
    \mathcal{L}_{\text{emb}} = \mathcal{L}_{\text{VAE}} + \mathcal{L}_{\text{WGAN}} + \lambda\mathcal{L}_{\text{CLS}}
    \label{L_embedding}
\end{equation}

Where $\lambda$ is a hyperparameter that balances the contribution of the classification loss with the semantic encoding loss, and the conditional representation.\newline

\textbf{(III) Representation.} Our last step involves training a conditional variational autoencoder (CVAE) \cite{mishra2018generative}. This is to ensure congruence between the latent representations extracted from the semantics and the learned distribution of samples within the true data manifold, Fig \ref{fig:architecture}(c). By learning a joint inference process $q_{\phi}(z_{seer};x,s)$, we can obtain a latent representation $z_{seer}$ while concurrently ensuring coverage of the true manifold and minimize biases towards the available training samples \cite{sohn2015learning}.
Furthermore, concatenating the synthetic visual data $G: z \rightarrow \hat{x}$ and semantic data space $s_C$ of the classes  $x'= x \oplus S$ means we can factorize the joint distribution into $p(x|z_{seer},s)p(z_{seer}|s)p(s)$. The encoder can encode more information about the true space through the variational posterior without losing class-specific information \cite{Beckham_2023}. To retain the distribution dependency of the true manifolds, the conditional mapping into the semantic encoded enhanced representation (SEER) space can be controlled with the regularisation parameter $\beta$. We assume a weak algebraic homomorphism of the mapping \cite{keurti2023homomorphism} when $\beta$ is low. The alignment of the learned representations can then be achieved by conditioning on the respective spaces with a reconstruction loss using our training data distribution \cite{wang2016relational}. Our loss function of the CVAE is then given as 

\vspace{-.3cm}
\begin{equation}
    \begin{split}
    \mathcal{L}_{\text{CVAE}}(l,k) = -\sum_{i=1}^{n} \mathbb{E}_{z_{seer} \sim q_{\phi}(z_{seer}|x'_i, s_i)} [ \log p_{\theta}(x'_i|s_i, z_i)] 
    \\ - \beta\ KL[q_{\phi}(z_i|s_i,x'_i) || p(z_i, s_i)]
    \end{split}
    %\mathcal{L}_{\text{CVAE}} = \mathcal{L}_{\text{recon}}(x', s) + \gamma \mathcal{L}_{\text{KL}}(\hat{z}, \mathcal{N}(\mu_z, \Sigma_z))
    \label{LCVAE}
\end{equation}

$\beta$ ensures that the manifested space $z_{seer}$ is retaining information from the probability distribution $p(\hat{z}|x')\sim\mathcal{N}(\mu_z, \Sigma_z)$ for spatial classification. This explicitly generated space will retain similar properties, even with reduced semantic quality in the training data. Specifically, the SEER provides properties of the probability space for a generalization over classes $\mathbb{E}_{z \sim \mathbb{P}_r}= ln[p_{\theta}(z|x',s)]  \forall i \in Y_{i}$ where $x' \in X'$. Assuming dependency of conditional prior $p(z_{seer},s^n)$ the KL term can be rewritten as \cite{Beckham_2023}

\vspace{-.3cm}
\begin{equation}
    \begin{split}
    \text{KL}(q_{\phi}(z_i|s_i,x'_i) || p(z_i, s_i))
    \\ = I_{\phi}(z;x',s) + KL(q_{\phi}(z)||(z|s) - \mathbb{E}_{q_{\phi}(s)}\log p(s)
    \end{split}
\end{equation}

where the last term is a constant and can therefore be ignored. Hence minimizing the evidence lower bound results in minimizing the mutual information between $z$ and $(x',s)$. As the semantic meanings are encoded in $x'$, we essentially want $z$ to encode independent factors in the data such that the true class information can be supervised by purely semantic information. However, reducing $I_{\phi}(z;s)$ necessarily also reduces $I_{\phi}(z;x')$. Therefore, the weighting of the KL term by $\beta$ allows us to exercise control over the factors in SEER, which is desirable when transforming unseen factors in ZSL.

\subsection{Final Classification}
For the final classification, we employ the cosine similarity in the SEER space. By assigning the positional label in the SEER space to the transformed test data, we aim to reduce class divergence. The surjective property ensures coverage of the manifold \cite{wang2016relational}. Furthermore, this approach enables the data to inherit the learned dependencies in the manifold structures, leveraging the modality invariant distribution, resulting in greater generalization. The class of a given sample $x \in X^u$ is determined by minimizing the cosine similarity in the SEER space:

\vspace{-.3cm}
\begin{equation}
\begin{split}
   \operatorname*{argmin}_y [\cos(q_{\phi}(seer^u|x^u, z^{all}),seer)] \\
    %= \{1 - \frac{\sum_{i}^{n}(q_{\phi}(z_i | x_i), z)}{z\sqrt{\sum_iz_i^2}} \}   \forall i \in Y
\end{split}
\end{equation}

\input{tables/table_dataset}
\input{tables/table_main_result}

%------------------------------------------------------------------------

%% file: tables/table_dataset.tex
\begin{table}[ht!]
\centering
\caption{Overview over benchmark dataset. Each datasets offer unique challenges and limitations for the model, resulting in a comprehensive understanding. Sample size refer to samples available per class. }
\scalebox{0.8}{
\begin{tabular}{@{}l|llll@{}}
\toprule
Dataset          & AWA2    & CUB    & FLO    & SUN    \\ \midrule
Sample size      & Large  & Small  & Medium & Small \\
Attribute size   & 85 & 312  & 64  & 1024 \\
Detail level     & Coarse & Fine   & Fine & Coarse \\
Class labels (S/U) & 40/10  & 150/50 & 82/20  & 645/72 \\ \bottomrule
\end{tabular}
}
\label{tabel:dataset}
\vspace{-.2cm}
\end{table}

%% file: tables/table_main_result.tex
\begin{table*}[hbt!]
\centering
\caption{\small{Performance of our proposed SEER-ZSL model in the generalized zero-shot setting over state-of-the-art generative ZSL models. Using the proposed split for unseen/seen classes per dataset, including 20\% seen classes in the generalized setting. The result is reported for seen (S), unseen (U), and harmonic mean (H= $[\frac{2*S*U}{S+U}]$). * indicates benchmark against AwA1 as AwA2 is not reported. \textcolor{red}{Red} and \textcolor{blue}{blue} marks best and second best results, respectively.}}
\scalebox{0.8}{
\begin{tblr}{
    rowsep=0pt,
  cells = {c},
  cell{1}{3} = {c=3}{},
  cell{1}{6} = {c=3}{},
  cell{1}{9} = {c=3}{},
  cell{1}{12} = {c=3}{},
  cell{3}{1} = {r},
  cell{4}{1} = {r},
  cell{5}{1} = {r},
  cell{6}{1} = {r},
  cell{7}{1} = {r},
  cell{8}{1} = {r},
  cell{9}{1} = {r},
  cell{10}{1} = {r},
  cell{11}{1} = {r},
  cell{12}{1} = {r},
  cell{13}{1} = {r},
  cell{14}{1} = {r},
  cell{15}{1} = {r},
  cell{16}{1} = {r},
  vline{3, 6, 9, 12} = {2}{},
  vline{2,3,6,9,12} = {1-16}{},
  hline{1,3,16} = {-}{},
  hline{2} = {3-14}{}
}
&   & \textbf{AWA2} & & & \textbf{CUB}  & & & \textbf{SUN}  & & & \textbf{FLO} & & \\
\textbf{Methods}  & \textbf{Venue} & \textit{S} & \textit{U} & \textit{H} & \textit{S} & \textit{U} & \textit{H} & \textit{S} & \textit{U} & \textit{H} & \textit{S} & \textit{U} & \textit{H} \\

%f-CLSWGAN* \cite{xian2018feature}          & IEEE'18       & 61.4 & 57.9 & 59.6 & 57.5 & 43.7 & 49.7 & -    & -    & 39.4 & 73.8 & 59.0 & 65.6 \\
f-VAEGAN-D2 \cite{xian2019f}               & CVPR'19       & 70.6 & 57.6 & 63.5 & 60.1 & 48.4 & 53.6 & 38.0 & 45.1 & 41.3 & 74.9 & 56.8 & 64.6 \\
LisGAN* \cite{li2019leveraging}            & CVPR'19       & 76.3 & 52.6 & 62.3 & 57.9 & 46.5 & 51.6 & 37.8 & 42.9 & 40.2 & \textcolor{blue}{83.8} & 57.7 & 68.3 \\
GDAN \cite{huang2019generative}            & ICCV'19       & 67.5 & 32.1 & 43.5 & \textcolor{blue}{66.7} & 39.3 & 49.5 & 40.9 & 38.1 & 53.4 & -    & -    & -\\
IZF \cite{shen2020invertible}              & ECCV'20       & 77.5 & 60.6 & 68.0 & \textcolor{red}{68.0} & 52.7 & 59.4 & \textcolor{blue}{57.0} & \textcolor{blue}{52.7} & \textcolor{blue}{54.8} & -     & -    & - \\
SDGZSL \cite{chen2021semantics}            & ICCV'21       & 73.6 & 64.6 & 68.8 & 66.4 & 59.9 & \textcolor{blue}{63.0} & -    & -    & -    & \textcolor{red}{90.2}  & \textcolor{red}{83.3} & \textcolor{red}{86.6}  \\
%DAA-GZSL \cite{zhao2023generating}  & TBD           & 79.9 & \textbf{65.7} & 72.1 & 65.5 & \textbf{66.1} & \textbf{65.8} & 38.7 & 47.8 & 42.8 & - & - & - \\
HSVA \cite{chen2021hsva}                   & NeurIPS'21    & 76.6 & 53.9 & 66.8 & 58.3 & 57.2 & 55.3 & 39.0 & 48.6 & 43.3 & -    & -    & -    \\
%ICCE \cite{kong2022compactness}            & CVPR'22       & \textcolor{blue}{82.3} & 65.3 & 72.8 & 65.5 & \textcolor{red}{67.3} & \textcolor{red}{66.4} & -    & -    & -    & \textcolor{blue}{86.5} & 66.1 & \textcolor{blue}{74.9} \\  
FREE+ESZSL\cite{cetin2022closedform}       & ICLR'22       & 78.0 & 51.3 & 61.8 & 60.4 & 51.6 & 55.7 & 36.5 & 48.2 & 41.5 & 82.2 & 65.6 & \textcolor{blue}{72.9} \\
TF-VAEGAN+ESZSL \cite{cetin2022closedform} & ICLR'22       & 74.7 & 55.2 & 63.5 & 63.3 & 51.1 & 56.6 & 39.7 & 44.0 & 41.7 & 83.2 & 63.5 & 72.1 \\
ZLAP \cite{ijcai2022p114}                  & IJCAI'22      & 76.3 & 74.7 & 75.5 & 32.4 & 25.5 & 28.5 & 48.1 & 47.2 & 47.7 & 68.2 & 54.7 & 60.7 \\
SE-GZSL \cite{kim2022semantic}             & AAAI'23       & 68.1 & 58.3 & 62.8 & 53.3 & 41.5 & 46.7 & 30.5 & 50.9 & 34.9 & -    & -    & -    \\
TPR \cite{chen2024tpr}                     & NeurIPS'24    & \textcolor{red}{87.1} & \textcolor{blue}{76.8} & \textcolor{red}{81.6} & 41.2 & 26.8 & 32.5 & 50.4 & 45.4 & 47.8 & 77.5 & 64.5 & 70.4 \\
MAIN \cite{verma2024meta}                  & WACV'24       & \textcolor{blue}{81.8} & 72.1 & 76.7 & 58.7 & \textcolor{blue}{65.9} & 62.1 & 40.0 & 50.1 & 48.8 & -    & -    & -    \\
\hline
SEER-ZSL                                   & (Ours)        & 78.8 & \textcolor{red}{77.9} & \textcolor{blue}{78.3} & 62.5 & \textcolor{red}{66.3} & \textcolor{red}{64.3} & \textcolor{red}{68.1} & \textcolor{red}{66.6} & \textcolor{red}{67.3} & 71.6 & \textcolor{blue}{68.0} & 69.8             
\end{tblr}
}
\label{table:main_res}
\end{table*}

%% file: sections/dataset.tex
\section{Dataset}
\label{sec:data}

As mentioned in Sec. \ref{small_data}, we validate our proposed approach by applying publicaly available datasets for zero-shot image classification; SUN Attribute (SUN) \cite{patterson2012sun}, CUB-200-2011 Bird (CUB) \cite{wah2011caltech}, Animals with Attributes (AWA2) \cite{xian2017zero}, and Oxford Flower dataset (FLO) \cite{nilsback2008automated}. The datasets varies in the number of samples, classes, semantic spaces, and split sizes, see Table \ref{tabel:dataset} for an overview. The SUN dataset contains the broadest range of semantic differences while the CUB dataset contains annotations with the finest granularity, including 312 attributes that indicate various visual traits. The FLO and AwA2 datasets contain more distinctive features in a coarser space. The visual features are extracted from the ResNet101 backbon \cite{he2016deep}, pretrained on ImageNet \cite{xian2018zero}.
To demonstrate the generalization potential of our model, we include three different semantic spaces. These include the original hand annotations (OG), textual extractions from word2vec (w2v) \cite{NIPS2013_9aa42b31}, visually grounded semantics annotations (VGSE) \cite{xu2022vgse}, and logic-based rules (LEN) \cite{dong2019neural} derived from the original space. These collections encompass different representations and therefore serve as an excellent proxy for real-world settings. We assume that the true joint distribution of the semantics and the visual is equal. Consequently, these datasets can be used to measure the model's capability of generating a concurrent SEER space, which directly correlates with its generalization ability.

\input{tables/table_semantic_space}

%% file: tables/table_semantic_space.tex
\begin{table*}[hbt!]
\centering
\caption{\small{Performance of the proposed SEER-ZSL model in generalized zero-shot learning and conventional unseen (conv.) settings across the semantic sources (Space). Results are reported using the standard splits for seen (S) and unseen (U) classes, along with the harmonic mean (H).  \textcolor{red}{Red} values indicate the best-performing model for each respective semantic source.}}
\scalebox{0.8}{
%\resizebox{\linewidth}{!}{%
\begin{tblr}{
    rowsep=0pt,
  row{even} = {c},
  row{5} = {c},
  row{7} = {c},
  row{9} = {c},
  column{1} = {r},
  cell{1}{3} = {c=4}{c},
  cell{1}{7} = {c=4}{c},
  cell{1}{11} = {c=4}{c},
  cell{2}{4} = {c=3}{},
  cell{2}{8} = {c=3}{},
  cell{2}{12} = {c=3}{},
  cell{3}{3} = {c},
  cell{3}{4} = {c},
  cell{3}{5} = {c},
  cell{3}{6} = {c},
  cell{3}{7} = {c},
  cell{3}{8} = {c},
  cell{3}{9} = {c},
  cell{3}{10} = {c},
  cell{3}{11} = {c},
  cell{3}{12} = {c},
  cell{3}{13} = {c},
  cell{3}{14} = {c},
  cell{4}{2} = {r},
  cell{5}{2} = {r},
  cell{6}{2} = {r},
  cell{7}{2} = {r},
  cell{8}{2} = {r},
  cell{9}{2} = {r},
  cell{10}{2} = {r},
  vline{2-4,5-8,11-12} = {1}{},
  vline{2-5,8-9,12} = {2}{},
  vline{2-4,7-8,11-12} = {2-10}{},
  hline{1,4,7,10-11} = {-}{},
  hline{2-3} = {3-14}{},
}
      &             & \textbf{AWA2}  &               &               &               & \textbf{CUB}  &            &               &            & \textbf{SUN} &               &               &               \\
Space & Method      & Conv.         & Gen.          &               &               & Conv.         & Gen.       &               &            & Conv.        & Gen.          &               &               \\
      &             & \textit{H}    & \textit{S}    & \textit{U}    & \textit{H}    & \textit{H}    & \textit{S} & \textit{U}    & \textit{H} & \textit{H}   & \textit{S}    & \textit{U}    & \textit{H}    \\
W2V   & f-VAEGAN-D2 & 58.4          & 59.0            & \textcolor{red}{46.7}          & \textcolor{red}{52.5}          & 32.7          & \textcolor{red}{44.5}     & \textcolor{red}{23.0}    & \textcolor{red}{30.3}    & 39.6         & 33.3          & 25.9          & 29.1          \\
      \cite{NIPS2013_9aa42b31} & CADA-VAE    & 49.0          & 60.1          & 38.6          & 47.0            & 22.5          & 39.7       & 16.3          & 23.1       & 37.8         & 28.2          & 26            & 27            \\
      & Ours        & 43.8          & \textcolor{red}{71.2} & 41.6          & \textcolor{red}{52.5}          & 15.5          & 19.3       & 20.1          & 19.7       & 35.3         & \textcolor{red}{44.6} & \textcolor{red}{33.2} & \textcolor{red}{38.1} \\
VGSE & f-VAEGAN-D2 & 61.3          & 66.7          & 45.7          & 54.2          & 35            & \textcolor{red}{45.7}       & 24.1          & \textcolor{red}{31.5}       & 41.1         & 35.7          & 25.5          & 29.8          \\
    \cite{xu2022vgse}  & CADA-VAE    & 52.7          & 61.6          & 46.9          & 53.9          & 24.8          & 44.5       & 18.3          & 25.9       & 40.3         & 29.6          & 29.4          & 29.5          \\
      & Ours        & 79.8 & \textcolor{red}{85.2} & \textcolor{red}{62.3} & \textcolor{red}{72.0} & 28.1 & 32.4       & \textcolor{red}{28.1} & 30.1       & 31.9         & \textcolor{red}{42.1} & \textcolor{red}{34.2} & \textcolor{red}{37.7} \\
LEN \cite{dong2019neural}   & Ours        & 78.2          & 89.3          & 56.7          & 69.4          & 64.4          & 86.1       & 33.18         & 47.9       & 41.6         & 52.4          & 52.2          & 52.3          
\end{tblr}
}
\label{tab:sem_space}
\end{table*}

%% file: sections/experiments.tex
\section{Experimental results}
\label{sec:perf_experiment}

We evaluate our approach following standard GZSL protocols, benchmarking top-1 accuracy against prior work in zero-shot classification. To assess the utility of our method, we analyze the contribution of each individual component, further supported by an ablation study providing additional evidence of the model’s effectiveness. \newline

%\begin{equation}
%\label{eq4}
%acc^{class}_{average}=\frac{1}{|Y|} \sum_{i=0}^{|Y|} \left(\frac{N_{correct_{class}}^{class_i}}%{N^{class_i}_{Total}}\right)
%\end{equation}

\textbf{Implementation details.} Our VAE and CVAE networks use a symmetric encoder-decoder MLP, compressing latent space $z$ to 48 dimensions for VAE and $d$ dimensions for SEER, where $d$ represents the semantic dimensions for the respective dataset. ReLU activation \cite{goodfellow2016deep} and a 0.1 dropout rate are employed for non-linearity and regularization. The WGAN's generator has three MLP layers (215, 516, 1024) and a 2048-dimensional output with a symmetric discriminator. To address the instability when sampling $z\sim \mathbb{P}_r$ we adopt tight training of the generator \cite{gulrajani2017improved}. The training schedule involves sequential iterations over the models, with partial integration at the end of each iteration. Furthermore, we use a learning rate of 0.001, exponential decay scheduling, and early stopping. The model, built in PyTorch, is trained on an NVIDIA RTX 8000 (48GB) for 3-5 hours, depending on the dataset size. \newline

%\begin{figure*}
%\centering
%    \begin{subfigure}[b]{0.2\textwidth}
%        \centering
%        \includegraphics[width=\textwidth]{figures/awa_lof_density.png}
%        \caption{ }
%    \end{subfigure}
%    \begin{subfigure}[b]{0.2\textwidth}
%        \centering
%        \includegraphics[width=\textwidth]{figures/sun_lof_density.png}
%        \caption{ }
%    \end{subfigure}
%    \begin{subfigure}[b]{0.2\textwidth}
%        \centering
%        \includegraphics[width=\textwidth]{figures/awa_isolation_forest_density.png}
%        \caption{ }
%    \end{subfigure}
%    \begin{subfigure}[b]{0.2\textwidth}
%        \centering
%        \includegraphics[width=\textwidth]{figures/sun_isolation_forest_density.png}
%        \caption{ }
%    \end{subfigure}
%    \label{fig:lof_density}
%    \caption{We measure the density of our SEER manifold to identify structural properties locally and globally. In a) and b)  we employ local outlier factor provides a relative density estimate for calculating varying density regions within classes. We see evenly distributed densities across semantic sources indicating similar manifold structures for the respective classes. For c) and d) we employ the global isolation forestry estimation, capturing density varation within classes.}
%\end{figure*}

\textbf{Generalized accuracy.} Table \ref{table:main_res} demonstrates the effectiveness of our SEER-ZSL model in achieving competitive or superior accuracy across multiple datasets in the generalized zero-shot learning (GZSL) setting. Specifically, our model surpasses the state-of-the-art in unseen class accuracy on three of the four benchmark datasets. For AWA2, CUB, and SUN our model outperforms in unseen accuracy by 11\%, 0.3\%, and 13.5\%, respectively. 
The SUN dataset’s diverse and granular class distribution challenges generalization, resulting in relatively lower performance across all models. However, our model achieves the highest harmonic mean (67.3\%) on SUN, showcasing its advanced semantic alignment and robust generalization capabilities. 
For the CUB dataset, SEER-ZSL achieves a marginal improvement in unseen class accuracy of 66.3\%, while trailing by only a 4.2\% drop in seen accuracy compared to the best-performing model. CUB is a fine-grained and attribute-rich dataset, making the balance of discriminative semantic properties particularly challenging. This challenge is evident in the performance of state-of-the-art models such as IZF \cite{shen2020invertible} and GDAN \cite{huang2019generative}, which, despite excelling in seen accuracy, suffer substantial declines of 27.4\% and 15.3\%, respectively, in unseen accuracy. In contrast, SEER-ZSL demonstrates a marginal unseen accuracy improvement of 3.8\%, highlighting the model’s focus on robust generalization.
In AWA, models like TRP \cite{chen2024tpr} and MAIN \cite{verma2024meta} tend to adapt well to the discriminative nature of the dataset with seen accuracies of 87.1\% and 81.8\%, but struggles to keep this performance across datasets. Our model accomplishes a second best harmonic mean of 78.3\% for and surpass the other models with 1.1\% for unseen classes.
FLO is characterized by a small semantic space with fine-grained details. Our model attains a second-best unseen accuracy of 68.0\%. 
%These results collectively showcase the adaptability and strength of SEER-ZSL in balancing seen and unseen class performance, achieving state-of-the-art or near state-of-the-art results across all the diverse benchmark datasets.

\begin{figure}
    \centering
    \begin{subfigure}{0.2\textwidth}
        \includegraphics[width=\textwidth]{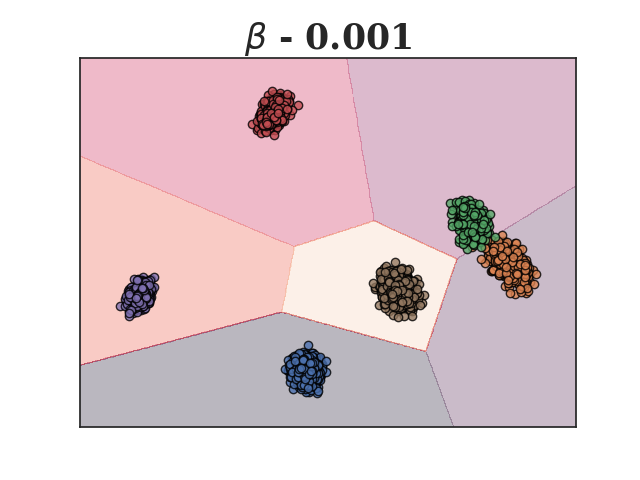}
        \caption{}
        \label{fig:latent_low}
    \end{subfigure}
    \begin{subfigure}{0.2\textwidth}
        \includegraphics[width=\textwidth]{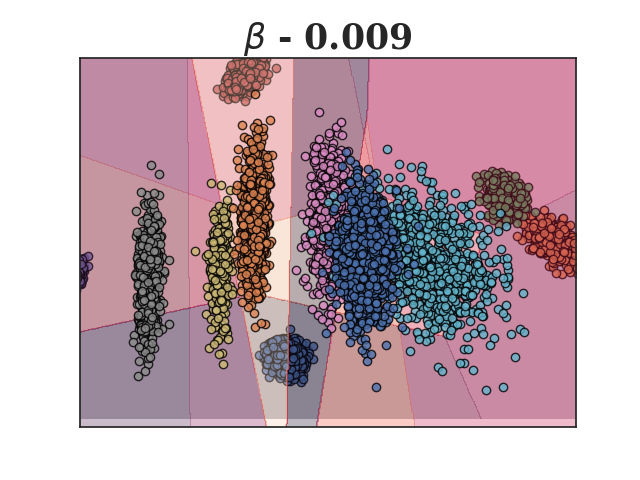}
        \caption{}
        \label{fig:latent_med}
    \end{subfigure}
    \begin{subfigure}{0.2\textwidth}
        \includegraphics[width=\textwidth]{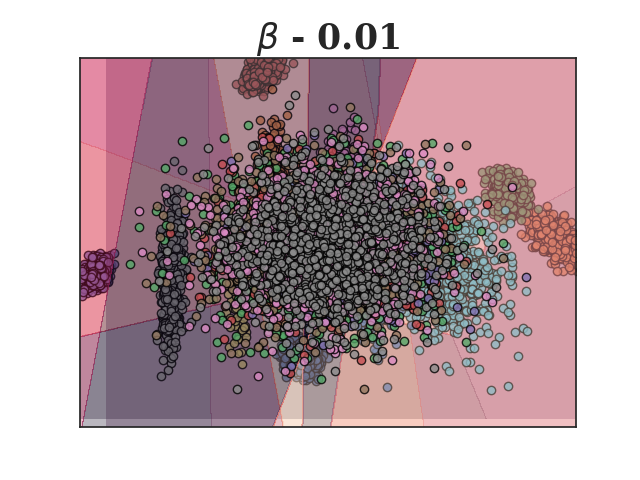}
        \caption{}
        \label{fig:latent_high}
    \end{subfigure}
    \begin{subfigure}{0.2\textwidth}
        \includegraphics[width=\textwidth]{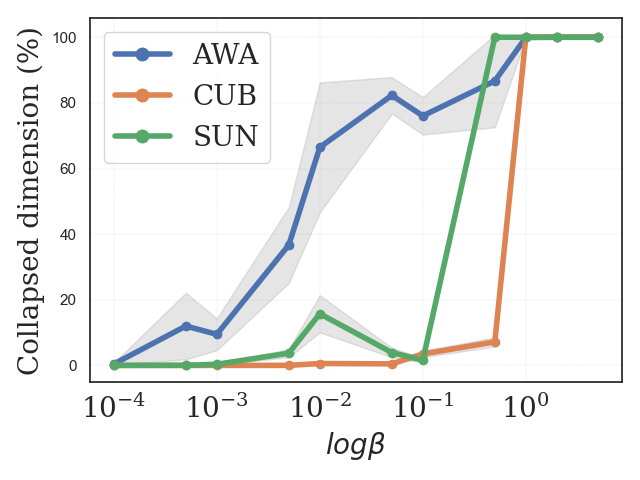}
        \caption{}
        \label{fig:latent_clp}
    \end{subfigure}
\caption{Control of the enhanced semantic latent space  $z$  for AWA2 is demonstrated under varying levels of the hyperparameter  $\beta$ in Eq. \ref{L_embedding}: (a) low $\beta$ , (b) medium  $\beta$ , and (c) high  $\beta$. Excessively strict regularization, as shown in subfigure (d), results in the collapse of the learned semantic space, highlighting the critical balance required for tuning  $\beta$.}
\label{fig:vae_beta}
\vspace{-.5cm}
\end{figure}

\begin{figure*}
    \centering
    \begin{subfigure}[b]{0.25\textwidth}
        \centering
        \includegraphics[width=\textwidth]{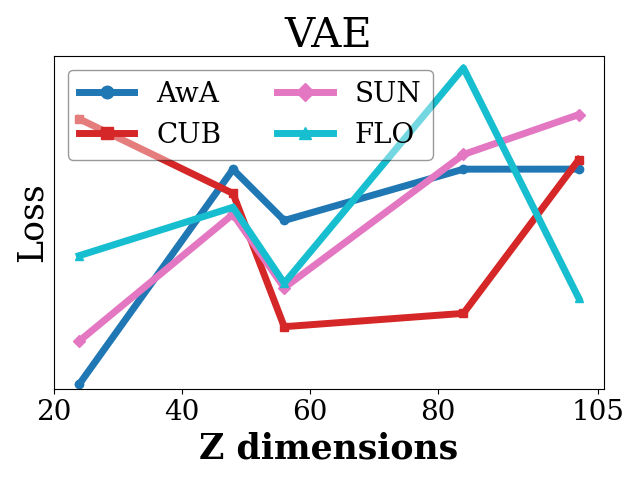}
        \caption{}
        \label{fig:z_dim_vae}
    \end{subfigure}
    \begin{subfigure}[b]{0.25\textwidth}
        \centering
        \includegraphics[width=\textwidth]{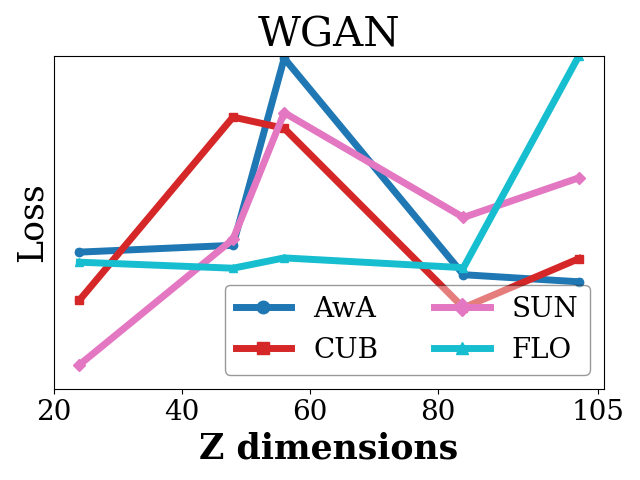}
        \caption{}
        \label{fig:z_dim_gan}
    \end{subfigure}
    \begin{subfigure}[b]{0.25\textwidth}
        \centering
        \includegraphics[width=\textwidth]{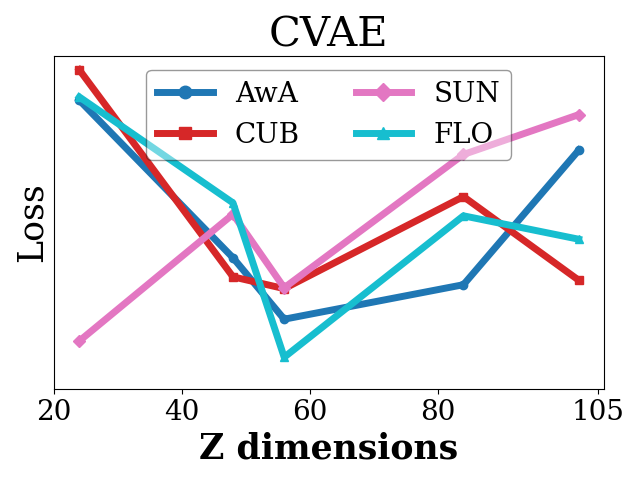}
        \caption{}
        \label{fig:z_dim_cvae}
    \end{subfigure}
    \caption{The impact of latent space dimensionality (z-dimensions) on the partial loss of the models is shown for (a) VAE, (b) WGAN, and (c) CVAE across all datasets (AWA2, CUB, SUN, and FLO). The plots highlight the trade-off between capturing discriminative features and ensuring sufficient coverage of the data manifold for unseen interpolation. The plots demonstrates an optimal balance at intermediate dimensions (48), achieving minimal loss for most datasets, emphasizing robust performance.}
    \label{fig:z_loss}
\end{figure*}

\begin{figure}
    \centering
    \includegraphics[width=0.9\linewidth]{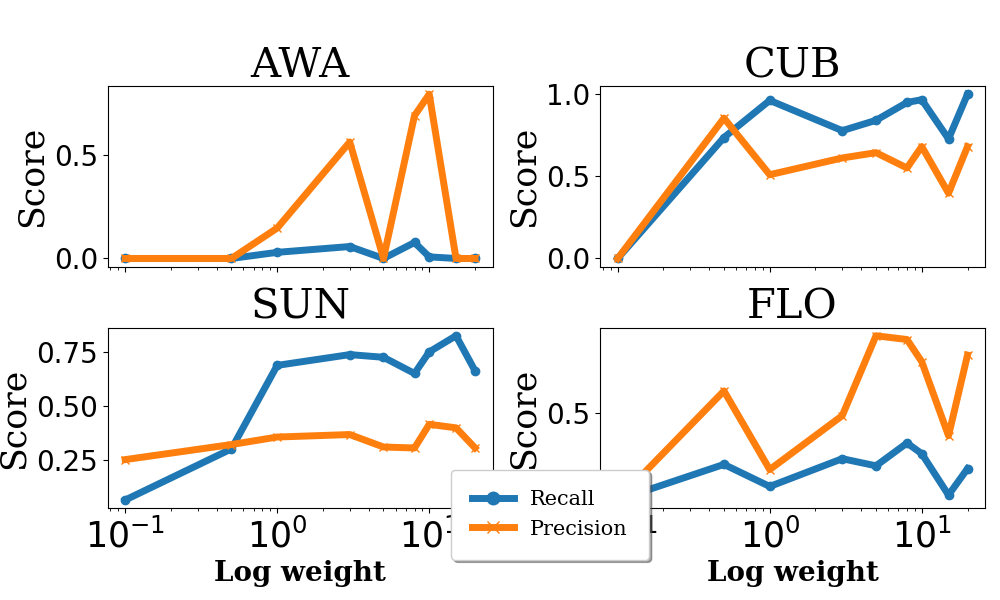}
    \caption{We use the metrics precision and recall as a proxy indicator for the transferability to unseen classes by evaluating the model’s ability to produce realistic and diverse samples, respectively. A high recall indicates that our model is able to cover the data manifold. By adjusting the weight of the classifier, $\lambda$, in Eq. \ref{L_embedding} we can attain a desirable balance.}
    \label{fig:recall_prec}
\vspace{-.5cm}
\end{figure}

\textbf{Semantic control.}
Table \ref{tab:sem_space} reports the performance of our model across three distinct semantic embeddings: W2V, VGSE, and LEN. As the semantic space serves as the primary source of information for bridging seen and unseen classes, evaluating performance across diverse semantic sources is essential to assess generalization and robustness. Using W2V embeddings, our model achieves harmonic means of 52.5\% on AWA and 38.1\% on SUN, matching state-of-the-art on AWA2 and surpassing prior methods by 9.0\% on SUN.
With visually grounded semantic embeddings (VGSE), our model further improves, outperforming state-of-the-art by 17.8\% on AWA2 and 7.9\% on SUN. LEN embeddings show the most significant gains, where our model achieves harmonic means of 69.4\% on AWA2, 47.9\% on CUB, and 37.7\% on SUN, consistently demonstrating superior performance across all datasets and embedding types.

We attribute these gains to the enhanced semantic control facilitated by our encoder architecture (Fig. \ref{fig:architecture}(a)), which enables more discriminative and structured latent representations. This improved semantic separability, illustrated in Fig. \ref{fig:vae_beta}, is achieved by tuning the hyperparameter  $\beta$  in Eq. \ref{LVAE}.
%ensuring optimal alignment of the visual and semantic spaces.
However, overly strong regularization can collapse the latent space towards the prior, as shown in Fig. \ref{fig:latent_clp}, hindering effective information transfer. Conversely, weak priors fail to capture meaningful interpolation for unseen classes, leading to suboptimal representations \cite{asperti2020balancing}. Hence, tuning $\beta$ enhances the robustness of our model across diverse semantic sources.

\begin{figure*}
    \centering
    \begin{subfigure}[b]{0.2\textwidth}
        \centering
        \includegraphics[width=\textwidth]{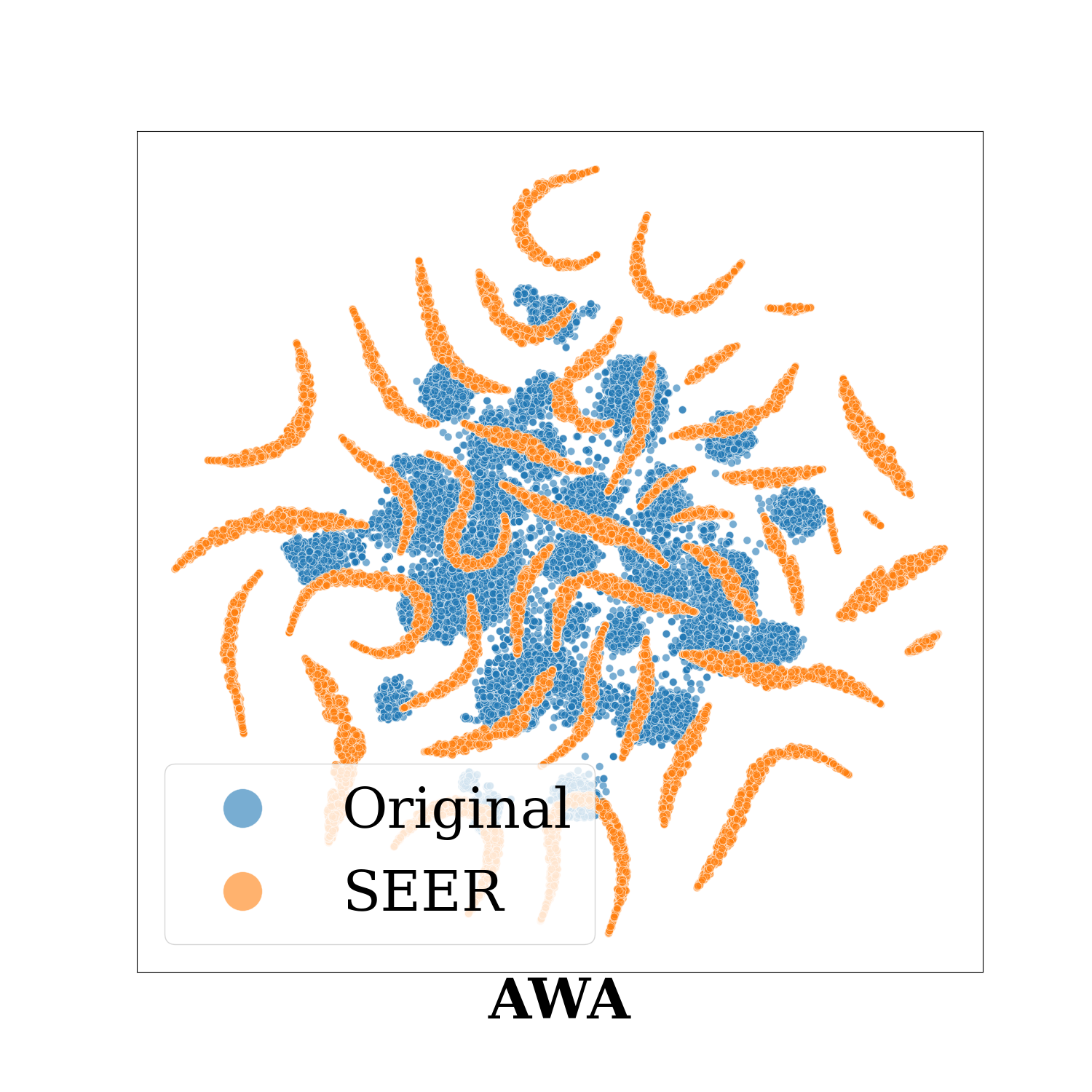}
        \caption{}
        \label{fig:tsne_subfig1}
    \end{subfigure}
    \hspace{-0.5cm}
    \begin{subfigure}[b]{0.2\textwidth}
        \centering
        \includegraphics[width=\textwidth]{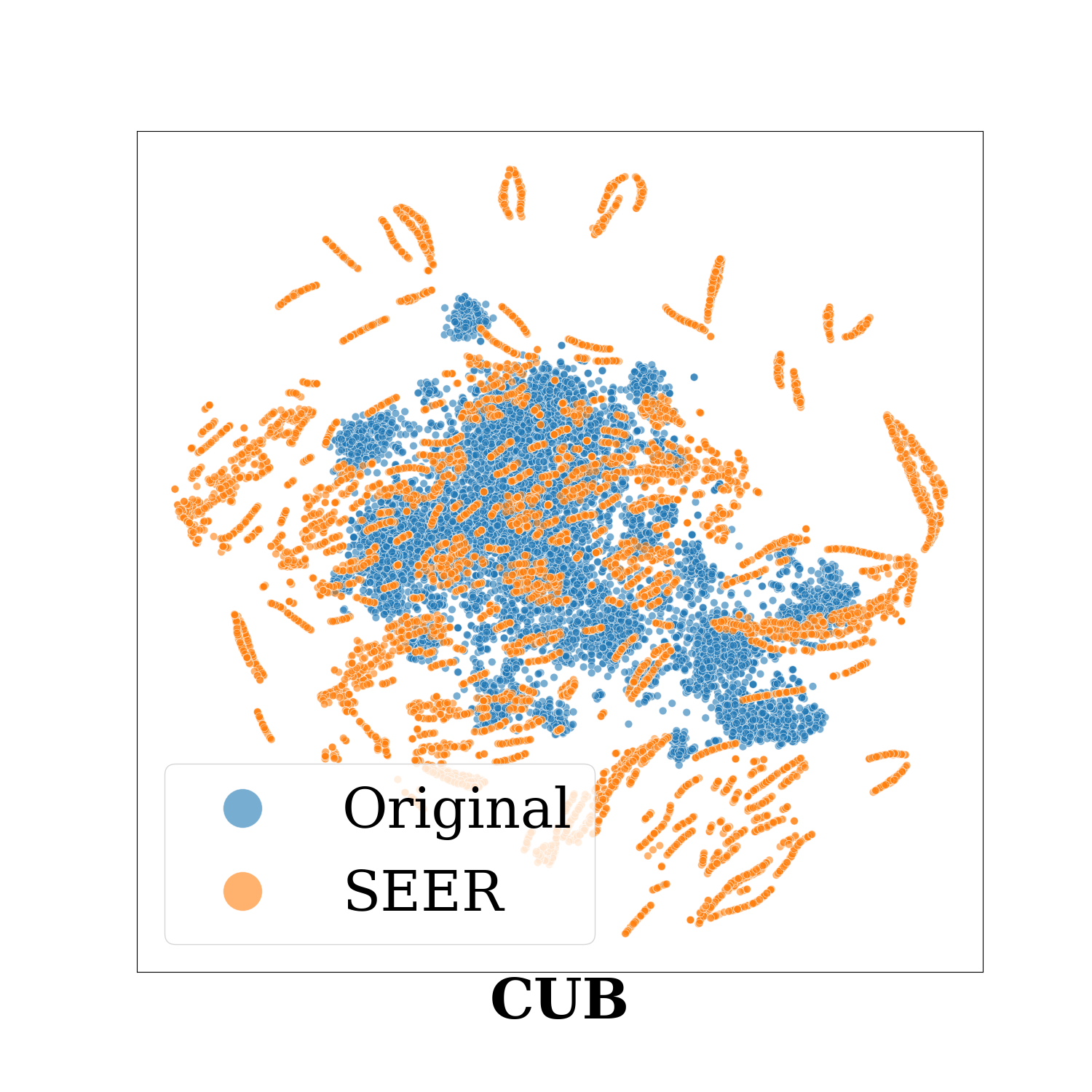}
        \caption{}
        \label{fig:tsne_subfig3}
    \end{subfigure}
    \hspace{-0.5cm}
    \begin{subfigure}[b]{0.2\textwidth}
        \centering
        \includegraphics[width=\textwidth]{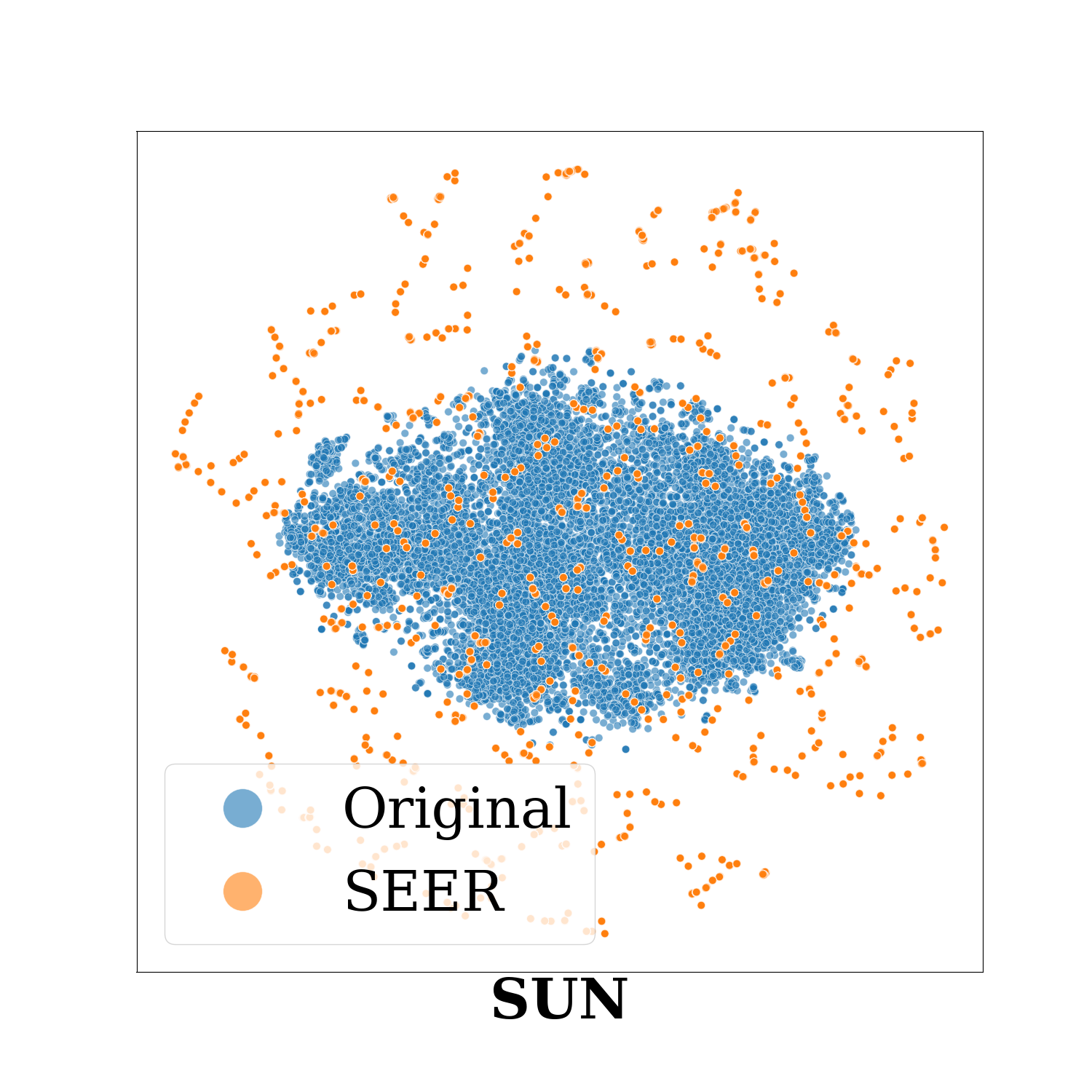}
        \caption{}
        \label{fig:tsne_subfig2}
    \end{subfigure}
    \hspace{-0.5cm}
%    \begin{subfigure}[b]{0.2\textwidth}
%        \centering
%        \includegraphics[width=\textwidth]{figures/apy_tsne.png}
%        \caption{}
%        \label{fig:tsne_subfig3}
%    \end{subfigure}
%    \hspace{-0.5cm}
    \begin{subfigure}[b]{0.2\textwidth}
        \centering
        \includegraphics[width=\textwidth]{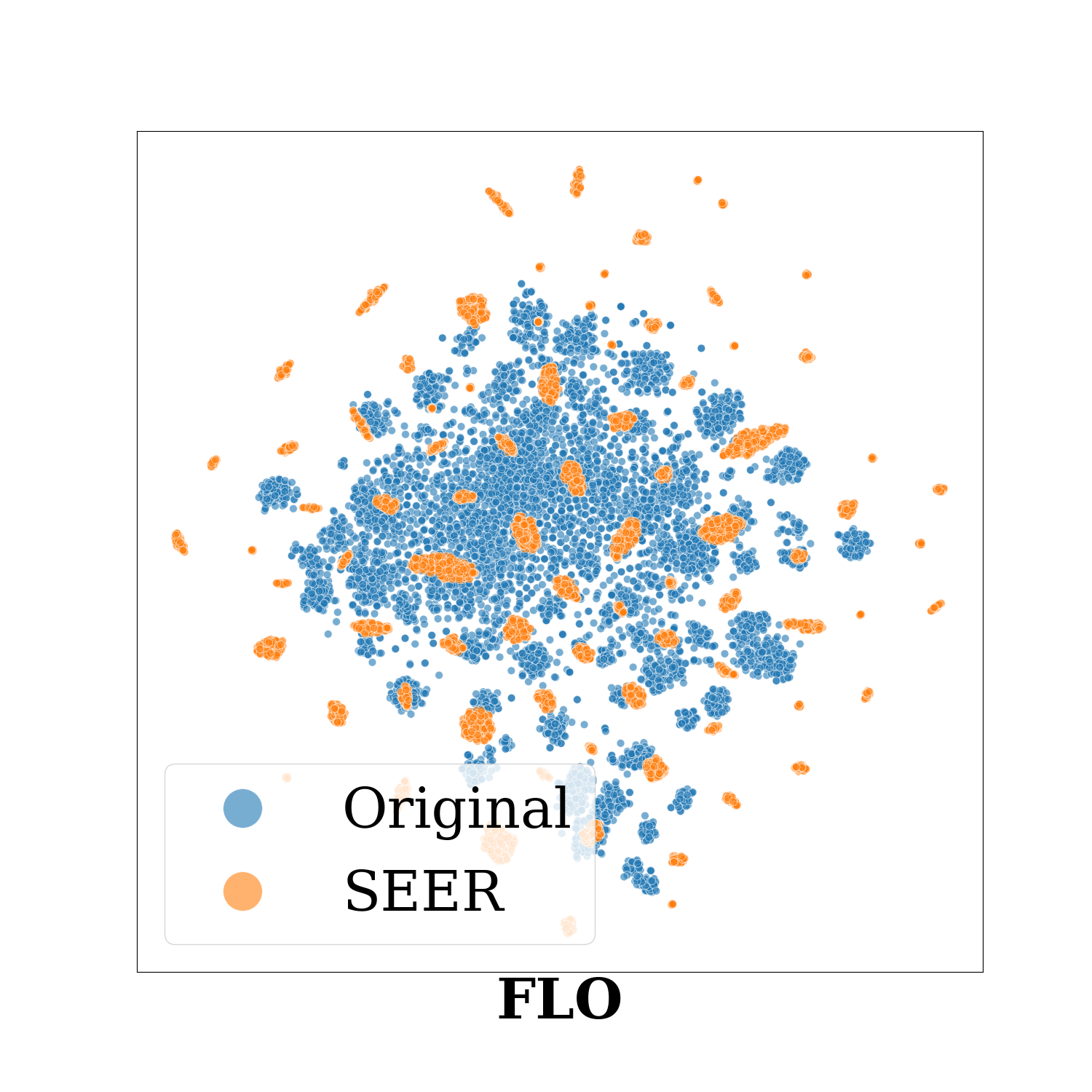}
        \caption{}
        \label{fig:tsne_subfig4}
    \end{subfigure}
    \caption{t-SNE \cite{van2008visualizing} of the original visual space (orange) compared to our SEER space (blue). SEER space suggests a greater distinctiveness and class separability, indicating that the model is better at preserving the local structure of the data in the semantic- and visual manifold. We also show a greater coverage of the distribution and the avoidance of local hubs.}
    \label{fig:tsne_seer}
\end{figure*}

Fig. \ref{fig:z_dim_vae} illustrates the effect of the semantic embedding dimensionality (z-dimension) on loss propagation for the models. The results demonstrate the critical trade-off between retaining relevant semantic information and achieving sufficient coverage of the data manifold for downstream tasks. Over-compression of the latent space risks discarding essential semantic features (Fig. \ref{fig:z_dim_cvae}), while overly large dimensions reduce the space’s discriminative capacity (Fig. \ref{fig:z_dim_vae}), impairing classification performance \cite{khrulkov2020hyperbolic}. Empirical evidence highlights that compressing the latent space to a dimension of 48 consistently yields optimal performance across all datasets (AWA2, CUB, SUN, and FLO), regardless of the dataset's true semantic complexity.
%This finding underscores the importance of carefully tuning the $z$-dimension to balance representation fidelity with model generalization.

\textbf{Coverage of manifold.} 
To transfer knowledge to unseen classes, the generator must align closely with the true probability distribution over the entire data manifold. This alignment is guided, in part, by the classification loss in Eq. \ref{L_embedding}, which ensures that the features generated maintain the discriminative properties of the visual data. 
Fig. \ref{fig:recall_prec} illustrates this alignment by analyzing precision and recall as proxies for data quality and manifold coverage, respectively
%, as proposed by Kynkäänniemi et al. 
\cite{kynkaanniemi2019improved}.
As the classification weight $\lambda$ increases, recall improves for semantically dense datasets such as CUB, SUN, and FLO, reflecting the coverage of the data manifold. In contrast, the effect is less pronounced for AWA, suggesting that the generator achieves reasonable manifold coverage through the enhanced semantic space ($z$) with lower $\lambda$. By dynamically adjusting the classification weight, we ensure fidelity to the underlying probability distribution while adequately covering the data manifold \cite{zhao2023generating}. \newline

\textbf{Analysis of the SEER Space.}
The discriminative properties of the SEER space are visualized using t-SNE projections \cite{van2008visualizing} in Fig. \ref{fig:tsne_seer}. This demonstrates the alignment of visual and semantic input spaces into the semantically enriched latent representation (SEER). This alignment captures the joint probability distribution of both modalities, allowing the model to interpolate effectively for unseen classes. When only visual information is available, the model leverages the stochastic latent space, with the semantic embeddings acting as a bridge for missing label information. Through semantic conditioning during training, the model learns to cluster visual inputs to consistent latent positions. 
%This enables accurate classification for unseen inputs.

%\textcolor{red}{For unseen classes, the SEER latent manifold can exhibit low-density regions due to the absence of direct visual training data \cite{suzuki2019learning}. However, our dual approach mitigates this challenge by structuring the latent space probabilistically, using semantic embeddings as anchors to guide interpolation. This property is validated in Fig. \ref{fig:lof_density}, where local and global density estimations reveal the robustness of the SEER manifold. In Fig. \ref{fig:lof_density} (a) and (b), local outlier factor (LOF) density estimation \cite{breunig2000lof} highlights the variations in local densities within the complex, heterogeneous manifold. The absence of extreme outliers (evident from the lack of left-tail anomalies) indicates that the SEER space maintains coherent local structures for both seen and unseen samples. Moreover, the consistent density distributions across different semantic sources confirm the uniformity of local structures within the manifold. On a global scale, isolation forest density estimation (IFOR) in Fig. \ref{fig:lof_density} (c) and (d) captures variations in density regions across semantic sources. While some alternations are observed between semantic sources, the density distributions remain sufficiently cohesive, ensuring robust manifold representation.}

\begin{figure}[h]
    \centering
    \begin{subfigure}[b]{0.2\textwidth}
        \centering
        \includegraphics[width=\textwidth]{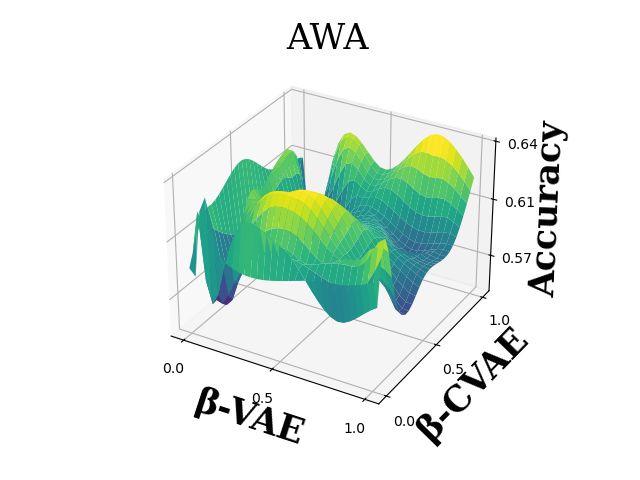}
        \caption{}
        \label{fig:subfig1}
    \end{subfigure}
    %\hfill
    \begin{subfigure}[b]{0.2\textwidth}
        \centering
        \includegraphics[width=\textwidth]{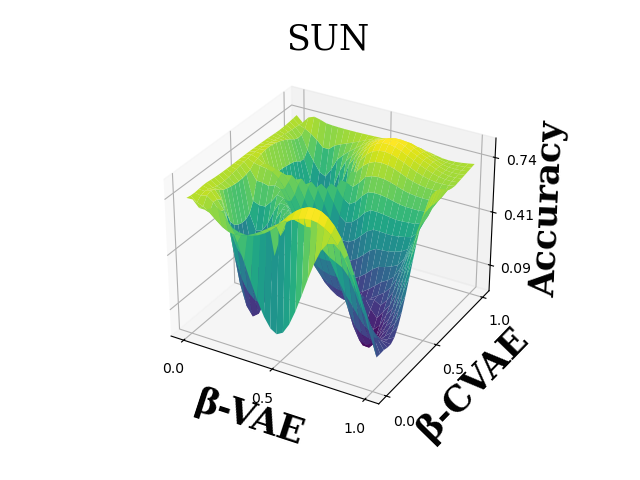}
        \caption{}
        \label{fig:subfig2}
    \end{subfigure}
    \caption{The semantic and visual information captured can be adjusted by the $\beta$-parameters in Eqs. \ref{LVAE} and \ref{LCVAE}. The figure shows measurements of unseen accuracy for (a) AWA and the semantically denser (b) SUN. This illustrates the complex landscape of aligning the VAE and CVAE.}
    \label{fig:3d_betas}
\end{figure}

The tunable hyperparameter $\beta$ in Eqs. \ref{L_embedding} and \ref{LCVAE} provides a mechanism for precise control over the SEER latent structure. Fig. \ref{fig:3d_betas} shows how this flexibility enables the model to adapt to both the discriminative AWA and the dense SUN dataset. By adjusting the $\beta$ values, the model balances the trade-off between semantic and visual influence, allowing for generalization across datasets.

\begin{table}[t]
\centering
\caption{Result from eliminating respective modules to evaluate the components capturing true data distribution on AWA2 with LEN semantics.}
    \scalebox{0.8}{
    \begin{tabular}{c|c|c}
    \hline
    $_{exc/}$ VAE & $_{exc/}$ WGAN & $_{exc/}$ CVAE \\ \hline
    23.8 & 33.9 & 40.0 \\ 
    \hline
    \end{tabular}
}
\label{tab:abo}
\vspace{-.2cm}
\end{table}

\textbf{Ablation study.} We perform an ablation study by constructing variants where key elements are excluded. This is achieved by replacing the dependent outputs of the removed components with isotropic Gaussian noise. The results, summarized in Table \ref{tab:abo}, highlight the critical role each component plays in achieving robust performance. Excluding the VAE results in a substantial performance drop to 23.8\%, nearly 40\% lower than the performance of the complete model. The VAE amplifies critical regions of the data manifold, and its absence hinders the model’s ability to capture a cohesive semantic structure. Removing the WGAN increases performance to 33.9\%, suggesting that while WGAN contributes to generative fidelity, it may not align as effectively with the semantic latent space in isolation. In contrast, excluding the CVAE leads to a significant performance degradation of 29\%, reflecting its central role in aligning the dual visual and semantic manifolds to create a cohesive latent representation. This highlights the importance of 
%of the training approach where all components work together to 
semantic alignment and manifold coverage.

%% file: sections/conclusion.tex
\section{Conclusion}
\label{sec:conclud}
In this paper, we introduced a novel method for zero-shot learning, addressing the limitations of state-of-the-art approaches in generalization to broader, more realistic settings. To overcome these challenges, we propose the Semantic Encoded Enhanced Representation (SEER), a robust and transformation-invariant classification space. Our experimental results demonstrate that SEER achieves superior classification accuracy across diverse datasets. Future work will focus on fully exploring the potential of this enhanced space. Key directions include improving generative capabilities to enable creative synthesis and novelty detection within the visual manifold. These efforts aim to expand the scope and applicability of ZSL, driving practical advancements in the field. \newpage